\documentclass[11pt]{article}

\usepackage[preprint]{acl}

\usepackage{times}
\usepackage{latexsym}

\usepackage[T1]{fontenc}

\usepackage[utf8]{inputenc}

\usepackage{microtype}

\usepackage{inconsolata}

\usepackage{graphicx}
\usepackage{amsmath}
\usepackage{mathtools}
\usepackage{booktabs}
\usepackage{multirow}
\usepackage{colortbl}
\usepackage{listings}
\usepackage{array}
\usepackage[T1]{fontenc}
\usepackage{tabularx}
\usepackage{multicol}
\usepackage{enumitem}

\definecolor{lightgray}{gray}{0.95}
\usepackage[most]{tcolorbox} 
\usepackage{xcolor}

\tcbset{
  mypromptstyle/.style={
    colback=green!5!white,
    colframe=green!50!black,
    coltitle=black,
    boxrule=0.7pt,
    arc=5pt,
    outer arc=2pt,
    left=10pt,
    right=10pt,
    top=8pt,
    bottom=8pt,
    fonttitle=\bfseries\sffamily,
    colbacktitle=yellow!15!white,
    enhanced,
    attach boxed title to top center={yshift=-2mm},
    boxed title style={
      colframe=green!50!black,
      colback=yellow!15!white,
      sharp corners
    },
    before upper={\ttfamily\small}
  }
}
\usepackage{hyperref}
\usepackage[all]{hypcap}

\usepackage{fancybox}
\usepackage{mdframed}
\usepackage{etoolbox}
\definecolor{lightgraybox}{gray}{0.96}  
\definecolor{grayborder}{gray}{0.7}
\newmdenv[
  topline=false,
  bottomline=false,
  rightline=false,
  leftline=true,
  linecolor=grayborder,
  linewidth=2pt,
  backgroundcolor=lightgraybox,
  roundcorner=4pt,
  innertopmargin=5pt,
  innerbottommargin=5pt,
  innerleftmargin=8pt,
  skipabove=4pt,
  skipbelow=4pt,
]{exampleblock}
\newif\ifcomment
\commenttrue
\ifcomment
    \newcommand{\shining}[1]{\textcolor{cyan}{ [{\em{shining}: \textcolor{cyan}{#1}}]}}
\else
  \newcommand{\shining}[1]{}
\fi

\title{Instructions for *ACL Proceedings}

\author{
\textbf{Jie Deng\textsuperscript{1}},
\textbf{Hanshuang Tong\textsuperscript{1}},
\textbf{Jun Li\textsuperscript{1}},
\textbf{Shining Liang\textsuperscript{1}},\\
\textbf{Ning Wu\textsuperscript{1}},
\textbf{Hongzhi Li\textsuperscript{1}},
\textbf{Yutao Xie\textsuperscript{1}}
\\[4pt]
\textsuperscript{1}Microsoft, Beijing, China
\\
\small{\texttt{dengjie530@outlook.com, junli1@microsoft.com}}
}

\begin{document}
\title{Beyond Rejection Sampling: Trajectory Fusion for Scaling Mathematical Reasoning}
\maketitle

\begin{abstract}
Large language models (LLMs) have made impressive strides in mathematical reasoning,
often fine-tuned using rejection sampling, which retains only correct reasoning trajectories.
While effective, this paradigm treats supervision as a binary filter that systematically excludes teacher-generated errors, leaving a gap in how reasoning failures are modeled during training.
\quad In this paper, we propose \textbf{TrajFusion}, a fine-tuning strategy that reframes rejection sampling as a structured supervision construction process. Specifically, TrajFusion forms fused trajectories that explicitly model trial-and-error reasoning by interleaving selected incorrect trajectories with reflection prompts and correct trajectories.
The length of the fused sample is adaptively controlled based on the frequency and diversity of teacher errors, providing richer supervision for challenging problems while safely reducing to vanilla rejection sampling fine-tuning (RFT) when error signals are uninformative. TrajFusion requires no changes to the architecture or training objective.
Extensive experiments across multiple math benchmarks demonstrate that TrajFusion consistently outperforms RFT, particularly on challenging and long-form reasoning problems.
\end{abstract}

\section{Introduction}

Large language models (LLMs) have made significant progress in mathematical reasoning, facilitated by techniques such as chain-of-thought prompting~\cite{Wei2022CoT} and supervised fine-tuning on high-quality solution trajectories~\cite{cobbe2021training,Yu2023MetaMathQA}. 
When combined with strong teacher models, these approaches enable LLMs to acquire increasingly sophisticated problem-solving behaviors and deliver exceptional accuracy across diverse mathematical benchmarks.

A widely adopted method for constructing supervised reasoning data is \emph{rejection sampling}~\cite{yuan2023scalingrelationshiplearningmathematical}. In this approach, a teacher model generates multiple candidate trajectories for a given problem, retaining only those with correct final answers. Rejection sampling fine-tuning (RFT) then performs supervised training on this filtered set of correct trajectories. RFT is appealing due to its simplicity, requiring no human annotation, and its demonstrated ability to improve model performance across a wide range of mathematical tasks~\cite{yuan2023rrhf,Tong2024DART,xiong2025minimalist}. Conceptually, RFT treats supervision as a binary filtering process, where correctness alone determines whether a trajectory contributes to learning.



However, this paradigm has a fundamental limitation: it systematically discards all
teacher-generated incorrect trajectories, despite the fact that such solutions contain
valuable diagnostic information about reasoning failures. Incorrect trajectories reveal
common computational slips, missing assumptions, misapplied formulas, and logically
plausible yet invalid argument paths. These failure patterns are not noise; rather, they
provide informative supervision that can help models recognize and avoid common reasoning
errors, especially on long or ambiguous problems where distinguishing valid from flawed
reasoning is essential~\cite{An2023LeMa,Pan2025LEMMA}. This leads to our central insight:
\emph{\textbf{teacher-generated errors should not be discarded, but incorporated as
structured supervision.}}


Building on this insight, we propose \textbf{TrajFusion}, a simple yet effective fine-tuning framework that extends RFT by incorporating selected incorrect reasoning trajectories into the supervision signal.
For each problem, TrajFusion constructs a single training sample by interleaving a selected subset of incorrect trajectories with the correct one, linked by brief reflection prompts. 
This approach explicitly simulates a trial-and-error reasoning process, exposing the model to both typical failure modes and their resolutions. 
The selection in TrajFusion is fully adaptive. When a problem yields diverse or informative errors, more incorrect trajectories are incorporated to enrich supervision; when error signals are sparse or uninformative, TrajFusion provably degenerates to vanilla RFT.
In our method, error diversity is treated as an explicit control signal rather than an implicit assumption, which allows the method to behave robustly across problems of varying difficulty.
Importantly, TrajFusion requires no additional sampling budget, no auxiliary models or critique generators, no architectural modifications, and no external difficulty labels.

We evaluate TrajFusion on two base models, LLaMA3-8B and DeepSeekMath-7B, across six mathematical reasoning benchmarks covering both in-domain and out-of-domain settings. Under both low-data (15K) and full-data (100K) regimes, TrajFusion consistently outperforms RFT, with particularly large gains on challenging datasets such as MATH, TheoremQA, and OlympiadBench. 
To assess its scalability, we further apply TrajFusion to long-form reasoning data distilled from DeepSeek-R1-Distill-Qwen-7B and fine-tune a 32K-context version of Qwen2.5-Math-7B. TrajFusion again surpasses RFT, achieving higher accuracy on competition-level benchmarks including AIME24, AIME25, and MATH-500.

\paragraph{Contributions.}
\begin{itemize}
    \item We formalize a previously underexplored supervision gap in RFT: the systematic exclusion of teacher-generated incorrect trajectories despite their diagnostic value for modeling reasoning failures. This motivates treating such errors as valuable supervision, rather than noise. 
    
    \item We propose \textbf{TrajFusion}, a structured supervision framework that reframes RFT from a binary filtering mechanism into a supervision construction process, adaptively fusing incorrect and correct reasoning trajectories through lightweight reflections. An adaptive mechanism automatically adjusts the amount of fused content based on the diversity and informativeness of teacher errors, while preserving standard RFT behavior when errors provide limited value.
    
    \item We demonstrate that TrajFusion consistently and scalably improves over RFT across multiple models, datasets, data regimes, and context lengths, with notable gains on challenging benchmarks and long-context mathematical reasoning.
\end{itemize}

\begin{figure*}[t]
    \centering
    \includegraphics[width=0.4\textwidth]{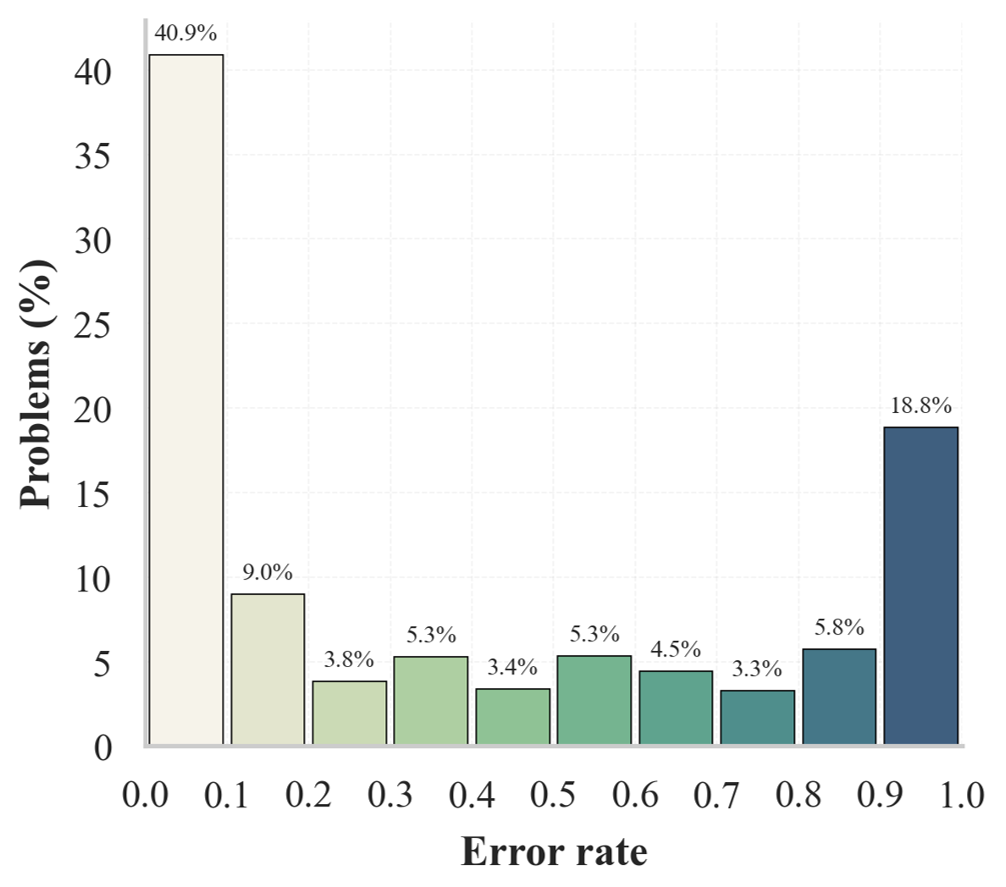}
    \hfill
    \includegraphics[width=0.5\textwidth]{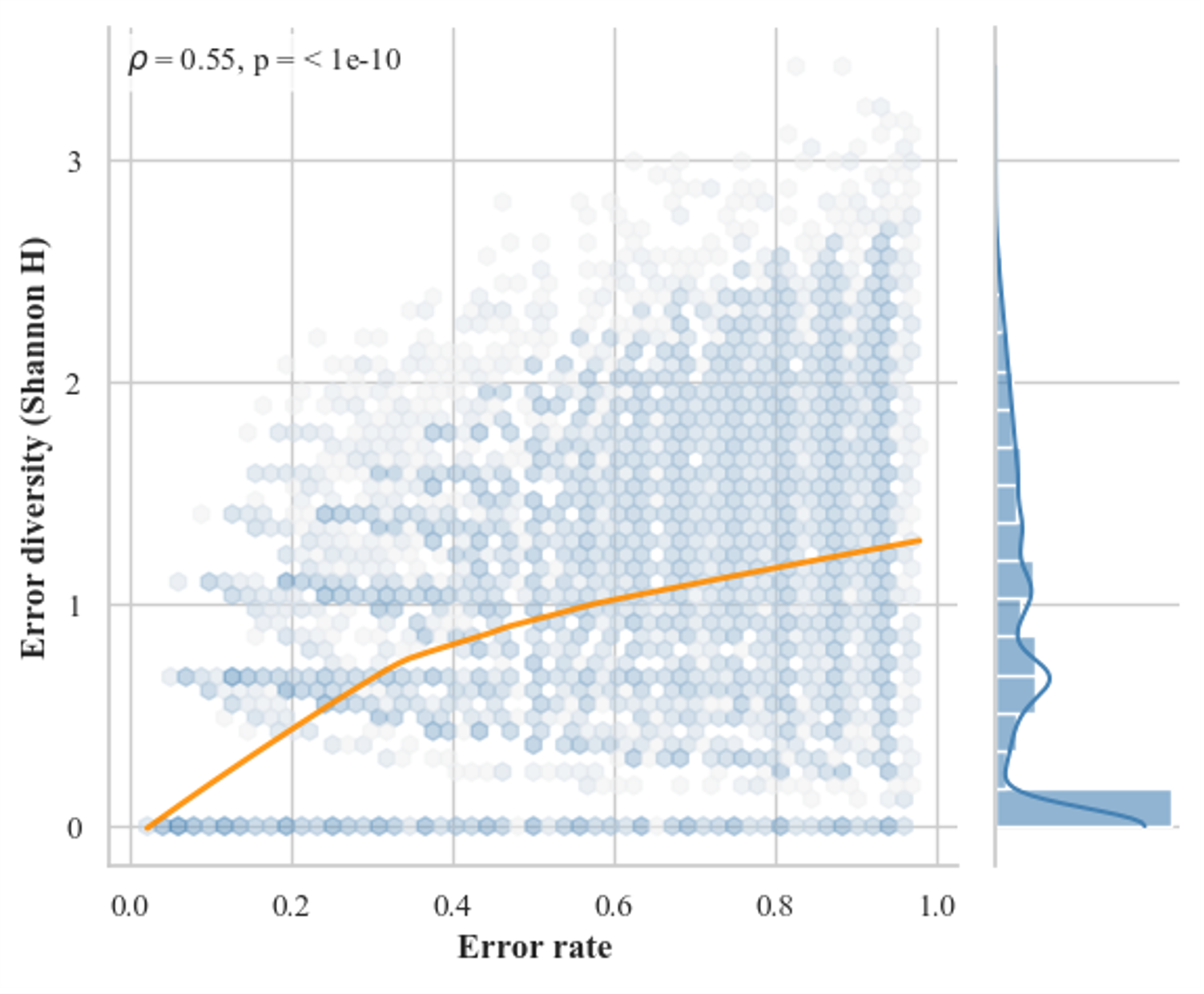}
    \caption{Sampling statistics for Qwen2.5-Math-7B-Instruct on the DeepMath dataset.
    Left: distribution of error rates across problems.
    Right: diversity of incorrect final answers measured by Shannon entropy}
    \label{fig:sampling-distributions}
\end{figure*}

\section{Preliminaries}
\label{sec:preliminary}

\subsection{Rejection Sampling Fine-Tuning}

Rejection sampling fine-tuning (RFT) constructs supervised data by sampling multiple
candidate trajectories for a problem $x$,
\[
\mathcal{Y}(x) = \{y_1,\dots,y_K\}, \qquad y_k \sim T(\cdot \mid x),
\]
and retaining only those with correct final answers,
\[
\mathcal{Y}^+(x)=\{y \in \mathcal{Y}(x) \mid \operatorname{corr}(y)=1\}.
\]
All incorrect trajectories,
\[
\mathcal{Y}^-(x)=\mathcal{Y}(x)\setminus\mathcal{Y}^+(x),
\]
are discarded. While this produces a clean training signal, it also removes the full
distribution of reasoning attempts. We analyze these sampling distributions to
understand what information RFT leaves unused.

\subsection{Sampling Distributions}

We analyze the sampling behavior of Qwen2.5-Math-7B-Instruct ~\cite{yang2024qwen2} on the DeepMath dataset~\cite{he2025deepmath} by examining error rates and the diversity of incorrect answers across sampled trajectories. Our analysis focuses on problems with mixed outcomes, where both correct and incorrect trajectories are observed.

Figure~\ref{fig:sampling-distributions} summarizes the sampling statistics.
The left panel shows the distribution of error rates across problems.
Approximately $40\%$ of problems have near-zero error rates and are almost always solved
correctly, while about $18\%$ are nearly always answered incorrectly.
Between these extremes lies a substantial set of problems with mixed outcomes, where the
model alternates between success and failure across attempts; these intermediate cases
are particularly informative for analyzing reasoning behavior under uncertainty.

The right panel characterizes failure structure by measuring the diversity of incorrect
final answers using Shannon entropy.
Error diversity increases with error rate ($\rho = 0.55$, $p < 10^{-10}$), indicating that
harder problems tend to induce a broader range of distinct failure modes rather than
repeated identical mistakes.

Beyond this overall trend, two characteristic patterns emerge. Some problems exhibit \textbf{low error rates but high diversity} among incorrect answers. In these cases, the model usually follows a reliable reasoning path, while rare failures arise from unstable or divergent intermediate steps. Other problems show \textbf{high error rates but low diversity}, where the model fails frequently yet converges to nearly the same incorrect answer each time. This behavior indicates a systematic misunderstanding or a strong inductive bias that repeatedly drives the model toward a specific incorrect solution.

Taken together, these observations demonstrate that the discarded set $\mathcal{Y}^-(x)$ encodes rich structural information beyond a binary notion of correctness. Incorrect trajectories vary not only in frequency but also in qualitative structure, reflecting distinct modes of reasoning instability and persistent misconceptions.

\section{Method}
\label{sec:method}

\begin{figure*}[t]
    \centering
    \includegraphics[width=\textwidth]{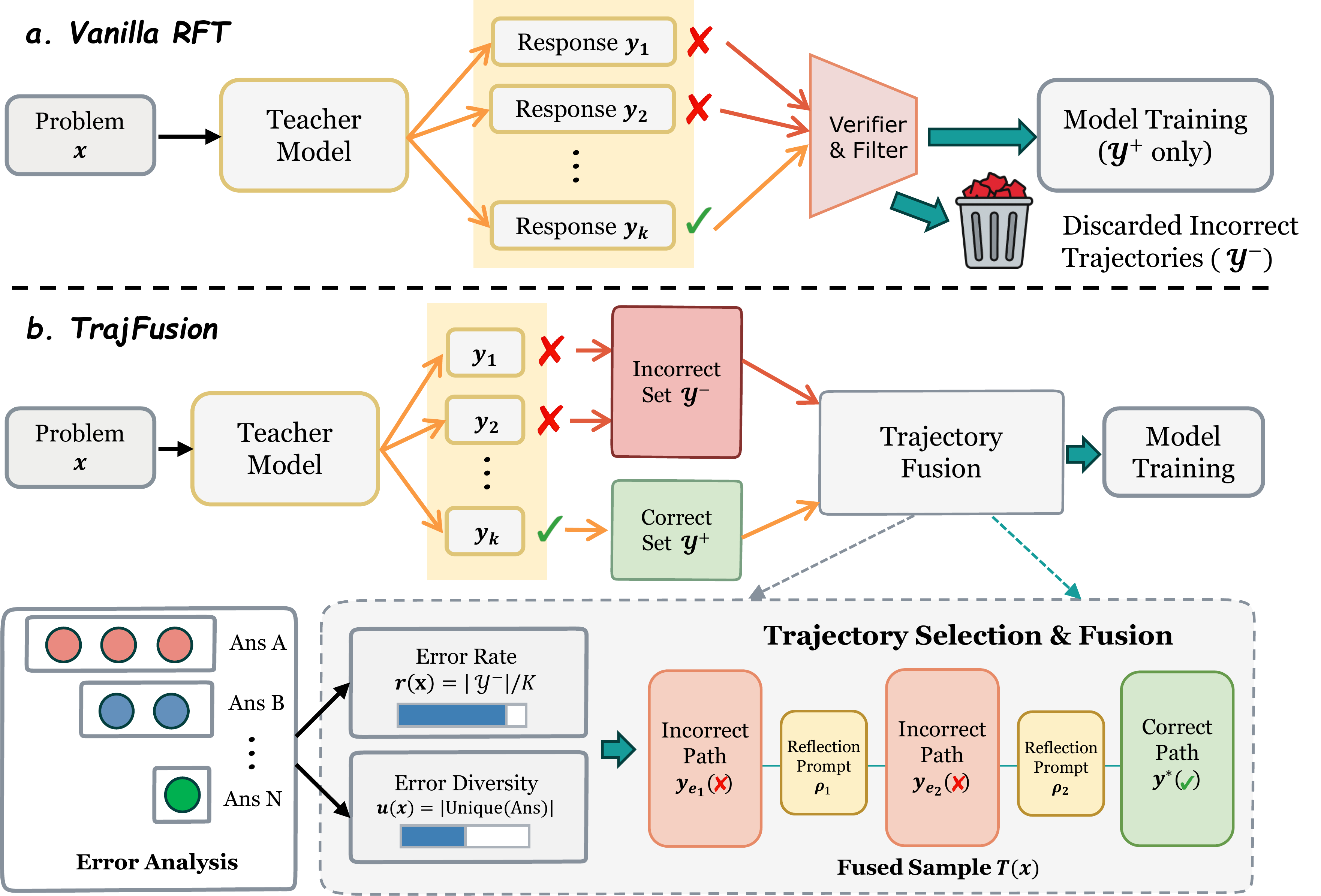}
    \caption{
    Comparison between vanilla rejection-sampling fine-tuning (RFT) and our TrajFusion framework.
    \textbf{Top:} Vanilla RFT samples multiple responses from a teacher model and retains only verified correct trajectories ($\mathcal{Y}^+$) for training, discarding all incorrect ones.
    \textbf{Bottom:} TrajFusion explicitly separates correct ($\mathcal{Y}^+$) and incorrect ($\mathcal{Y}^-$) trajectories, performs problem-level error analysis, and constructs fused training trajectories by selectively integrating informative incorrect reasoning paths with corrected trajectories.
    }
    \label{fig:compared}
\end{figure*}

Building on the observations in Section~\ref{sec:preliminary}, we introduce \textbf{TrajFusion}, a structured fine-tuning approach that integrates selected incorrect reasoning trajectories together with a correct one into a single fused training sample. 
Instead of relying exclusively on the correct solution, TrajFusion leverages information from diverse failure modes to provide richer supervision, while remaining fully compatible with vanilla RFT pipelines. 

\subsection{Sampling and Scoring}

For each problem $x$, we sample $K$ reasoning trajectories
\[
\mathcal{Y}(x) = \{ y_1, \dots, y_K \}, \qquad y_k \sim T(\cdot \mid x),
\]
and partition them into correct and incorrect subsets using an automatic verifier,
yielding $\mathcal{Y}^+(x)$ and $\mathcal{Y}^-(x)$ respectively.

To characterize the sampling outcomes at the problem level, we compute two statistics. The first is an
error-rate estimate
\[
r(x) = \frac{|\mathcal{Y}^-(x)|}{K},
\]
which measures how frequently the teacher model produces incorrect trajectories among the sample attempts for problem $x$.

The second statistic quantifies the diversity of incorrect outcomes:
\[
u(x) = \left|\,\{\operatorname{Ans}(y) : y \in \mathcal{Y}^-(x)\}\,\right|.
\]
Here distinct final answers represent different error types, allowing us to distinguish
between problems where the model collapses to a single failure mode and those where it
explores multiple incorrect reasoning paths.

\subsection{Difficulty- and Diversity-Aware Selection}

TrajFusion selectively incorporates incorrect trajectories into training.
The selection is jointly guided by error frequency and diversity: incorrect trajectories are included only when they occur with sufficient frequency, and a larger number of incorrect trajectories are selected as failures become more diverse. 

The number of retained incorrect trajectories is determined by
\[
k(x)=\min\!\left(k_{\max},\ \left\lfloor \alpha\, r(x)\, u(x) \right\rfloor\right),
\]
where $k_{\max}$ is a fixed upper bound and $\alpha$ controls the sensitivity to error diversity. This rule yields an adaptive behavior:
\begin{itemize}
    \item When no incorrect trajectory is observed,
$r(x)=0$ and thus $k(x)=0$, reducing TrajFusion to vanilla RFT without introducing any
additional error supervision.
    \item When errors are present, $k(x)$ increases with both the error rate $r(x)$ and the answer diversity $u(x)$: homogeneous failures (small $u(x)$) result in fewer selected incorrect trajectories, while higher error rates and more diverse outcomes lead to richer supervision.
\end{itemize}

To obtain representative yet compact supervision, we organize incorrect trajectories $\mathcal{Y}^-(x)$ by their final answers and prioritize error groups by frequency. Within each selected group, we favor shorter trajectories as representatives, encouraging concise supervision while preserving diversity across failure outcomes.

\subsection{Fused Sample Construction}

Given a set of selected incorrect trajectories $\{y_{e_1}, \dots, y_{e_k}\}$ and a correct trajectory $y^\star \in \mathcal{Y}^+(x)$, TrajFusion constructs a single fused training sample
\[
T(x) = [\,y_{e_1},\ \rho_1,\ \dots,\ y_{e_k},\ \rho_k,\ y^\star \,],
\]
where each $\rho_i$ is a fixed reflection prompt that explicitly marks the transition between successive attempts and provides contextual signals for correction.

The training of TrajFusion proceeds with standard supervised fine-tuning on the fused sequence sample. The model is optimized via next-token prediction using the negative log-likelihood objective
\[
\mathcal{L}_{SFT} = - \sum_{t=1}^{|T(x)|} \log p_\theta(T_t \mid T_{< t}),
\]
without any modification to the model architecture or training objective.
TrajFusion therefore operates as a supervision construction mechanism, that enriches the learning signal while maintaining compatibility with existing fine-tuning pipelines.

\begin{table*}[!t]
\centering
\small
\caption{
\textbf{Pass@1 accuracy on six mathematical reasoning benchmarks.}
Bold numbers indicate the best performance under the same base model.
TrajFusion consistently outperforms vanilla RFT and prior data construction methods across
both model families and data scales.
}
\label{tab:baseline-results}
\setlength{\tabcolsep}{9.5pt}
\renewcommand{\arraystretch}{1.25}
\begin{tabular}{lccccccc|c}
\toprule
\multirow{2}{*}{\textbf{Method}} &
\multirow{2}{*}{\textbf{Train Size}} &
\multicolumn{6}{c}{\textbf{Mathematical Reasoning Benchmarks}} &
\multirow{2}{*}{\textbf{AVG}} \\
\cmidrule(lr){3-8}
& & \textbf{MATH} & \textbf{GSM8K} & \textbf{College}
& \textbf{DM} & \textbf{Olympiad} & \textbf{Theorem} & \\
\midrule

\multicolumn{9}{c}{\textbf{\textit{LLaMA3-8B Base Model}}} \\
\midrule

ICL                     & —     & 21.2 & 51.0 & 19.9 & 27.4 & 4.2  & 19.8 & 23.9 \\
MetaMath                & 400k  & 32.5 & 77.3 & 20.6 & 35.0 & 5.5  & 13.8 & 30.8 \\
MMIQC                   & 2.3M  & 39.5 & 77.6 & 29.5 & 41.0 & 9.6  & 16.2 & 35.6 \\
DARTMath                & 590k  & 46.6 & \textbf{81.1} & 28.8 & \textbf{48.0} & 14.5 & 19.4 & \underline{39.7} \\
MathFusion               & 60k   & 46.5 & \underline{79.2} & 27.9 & 43.4 & \underline{17.2} & 20.0 & 39.0 \\
LEMMA                   & 89k   & 38.3 & \underline{79.2} & 24.7 & —    & —    & —    & — \\
\rowcolor{gray!10}
Vanilla RFT             & 15k   & 29.5 & 72.3 & 15.8 & 28.4 & 6.4  & 14.9 & 27.9 \\
\rowcolor{gray!10}
             & 100k  & \underline{48.5} & 76.9 & \underline{28.9} & 41.1 & 14.5 & \underline{22.6} & 38.8 \\

\rowcolor{blue!10}
\textbf{TrajFusion} & 15k   & 36.6 & \underline{79.2} & 21.6
                          & 37.9 & 8.6 & 20.0 & 34.0 \\
\rowcolor{blue!10}
 & 100k  & \textbf{51.4} & 77.3 & \textbf{31.6}
                          & \underline{43.7} & \textbf{20.2} & \textbf{23.3} & \textbf{41.3} \\
\midrule

\multicolumn{9}{c}{\textbf{\textit{DeepSeekMath-7B Base Model}}} \\
\midrule

ICL                     & —     & 35.5 & 64.2 & 34.7 & 45.2 & 9.3  & 23.5 & 35.4 \\
Instruct                & 780k  & 46.9 & \underline{82.7} & 37.1 & 52.2 & 14.2 & 28.1 & 43.5 \\
MMIQC                   & 2.3M  & 45.3 & 79.0 & 35.3 & 52.9 & 13.0 & 23.4 & 41.5 \\
DARTMath                & 590k  & 53.6 & \textbf{86.8} & \underline{40.7} & 61.6 & 21.7 & 32.2 & 49.4 \\
MathFusion               & 60k   & 53.4 & 77.9 & 39.8 & 65.8 & 23.3 & 24.6 & 47.5 \\
LEMMA                   & 89k   & 50.6 & 80.4 & 35.6 & —    & —    & —    & — \\

\rowcolor{gray!10}
Vanilla RFT             & 15k   & 51.8 & 78.5 & 38.3 & 64.9 & 19.6 & 21.4 & 45.8 \\
\rowcolor{gray!10}
             & 100k  & \underline{56.2} & 79.2 & 40.1 & \underline{69.1} & \underline{25.6} & 31.2 & \underline{50.2} \\

\rowcolor{blue!10}
\textbf{TrajFusion} & 15k   & 55.8 & 81.4 & 39.0
                          & 66.8 & 23.1 & \textbf{26.9} & 48.8 \\
\rowcolor{blue!10}
 & 100k  & \textbf{59.1} & 82.5 & \textbf{41.2}
                          & \textbf{69.6} & \textbf{27.0} & \textbf{34.1} & \textbf{52.3} \\
\bottomrule
\end{tabular}
\end{table*}

\section{Experiments}

\subsection{Implementation Details}

\paragraph{Dataset Construction and Sampling.}
We construct two training sets: an initial 15K set and an expanded 100K set. The 15K set
is drawn from GSM8K~\cite{cobbe2021training}, MATH~\cite{hendrycks2021measuring}, and
AIME (pre-2024). To introduce additional variety, we incorporate more problems from the
DeepMath dataset~\cite{he2025deepmath}, forming the 100K set.

For each problem $x$, we sample $K=16$ chain-of-thought trajectories using
Qwen2.5-Math-7B-Instruct~\cite{yang2024qwen2} with temperature 0.7 and top-$p$ 0.95,
generated through vLLM~\cite{kwon2023efficient} with a 4096-token limit. An automatic
verifier checks final answers, and queries without any correct solution are excluded
from training.

\paragraph{Training Setup.}
We fine-tune two base models, LLaMA3-8B-base~\cite{dubey2024llama} and
DeepSeek-Math-7B-base~\cite{shao2024deepseekmath} using ms-swift framework~\cite{zhao2024swiftascalablelightweightinfrastructure}. Training uses a batch size of 64,
learning rate $5\times10^{-5}$, and sequence packing with ZeRO-1 optimization
\cite{rasley2020deepspeed}. The 15K set is trained for 5 epochs and the 100K set for 3
epochs.

\paragraph{Evaluation Benchmarks.}
We evaluate on a suite of mathematical reasoning benchmarks, including
GSM8K~\cite{cobbe2021training}, MATH~\cite{hendrycks2021measuring},
CollegeMath~\cite{tang2024mathscale}, DeepMind-Mathematics~\cite{saxton2018analysing},
OlympiadBench-Math~\cite{he2024olympiadbench}, and TheoremQA~\cite{chen2023theoremqa}.
All scores report Pass@1 accuracy.

\paragraph{Baselines.}
We compare TrajFusion with representative data construction and fine-tuning strategies
for mathematical reasoning, including MMIQC~\cite{liu2025augmenting},
DART-Math~\cite{Tong2024DART}, MathFusion~\cite{pei2025mathfusion},
DeepSeekMath-7B-Instruct~\cite{shao2024deepseekmath}, and LEMMA~\cite{Pan2025LEMMA}.
As our primary controlled baseline, we include \textit{vanilla RFT}, which fine-tunes only on the correct trajectories obtained from rejection sampling.
This allows us to isolate the contribution of incorporating selected incorrect reasoning paths.

\subsection{Main Results}

\paragraph{Overall performance and data efficiency.}
Table~\ref{tab:baseline-results} summarizes the performance of TrajFusion across six
mathematical reasoning benchmarks. With only 15k training examples, TrajFusion already
achieves competitive results on both backbones. On DeepSeekMath-7B, it reaches 55.8\% on
MATH and 48.8\% on average, outperforming several data-construction baselines that use
substantially larger training sets (e.g., MathFusion at 60k and MMIQC at 2.3M). This
demonstrates that TrajFusion can extract meaningful supervisory signals even in low-data
regimes.

At 100k examples, TrajFusion achieves the highest average accuracy among all compared
training strategies: 41.3\% on LLaMA3-8B and 52.3\% on DeepSeekMath-7B. The method
continues to yield improvements as data scales, indicating that trajectory-level
supervision from both correct and selected incorrect attempts remains beneficial in
larger training regimes.

\paragraph{Effectiveness varies with reasoning complexity.}
TrajFusion provides the largest gains on benchmarks characterized by longer or more
structured reasoning, such as MATH, OlympiadBench, and TheoremQA. These tasks naturally
produce more diverse incorrect attempts, aligning with the model's strength in leveraging
trajectory-level variation. At 100k examples, TrajFusion improves over DARTMath by +4.8
points on MATH for LLaMA3-8B and +5.5 points for DeepSeekMath-7B, and achieves the highest
accuracy on TheoremQA across all methods.

On GSM8K, TrajFusion performs similarly to other approaches, with smaller gains than on
the more complex benchmarks. GSM8K problems involve short, well-structured reasoning
chains with limited variability, resulting in fewer opportunities for trajectory fusion
to provide additional benefit. 

\paragraph{Comparison with vanilla RFT.}
To isolate the contribution of integrating incorrect trajectories, the vanilla RFT
baseline uses the same sampled queries and the same correct trajectories as TrajFusion, but
discards all incorrect attempts. Across both model backbones and training sizes,
TrajFusion consistently outperforms vanilla RFT. For LLaMA3-8B, the improvements are
+6.1 points at 15k and approximately +3 points at 100k. For DeepSeekMath-7B, the gains
are +3.0 and +2.1 points respectively.

Gains are especially pronounced on structurally challenging benchmarks. On OlympiadBench,
TrajFusion improves LLaMA3-8B from 6.4\% to 8.6\%, and DeepSeekMath-7B from 19.6\% to
23.1\% at 15k. On TheoremQA, improvements of 5--6 points are observed on both backbones.
These results support the central hypothesis of TrajFusion: selectively incorporating
informative incorrect reasoning trajectories provides richer supervision than
correctness-only training, yielding consistent benefits in both low- and high-data
settings.

\begin{table}[t]
\centering
\caption{
Ablation on the effect of fusion count ($k_{\max}$) and error diversity on
\textbf{GSM8K} and \textbf{MATH} using \textbf{LLaMA3-8B}.
\#Err denotes the average number of fused incorrect trajectories per example.
}
\label{tab:combined-fusion-diversity}
\small
\setlength{\tabcolsep}{6pt}
\renewcommand{\arraystretch}{1.15}
\begin{tabular}{@{}l c rr rr@{}}
\toprule
\multirow{2}{*}{\textbf{Strategy}} &
\multirow{2}{*}{\textbf{\#Err}} &
\multicolumn{2}{c}{\textbf{GSM8K}} &
\multicolumn{2}{c}{\textbf{MATH}} \\
& & Acc & Tokens & Acc & Tokens \\
\midrule
TF($k_{\max}{=}1$)          & 0.62 & 72.0 & 125  & 39.6 & 912  \\
TF($k_{\max}{=}2$)          & 1.02 & 72.9 & 145  & 41.9 & 1056 \\
\rowcolor{gray!8}
\textbf{TF($k_{\max}{=}4$)} & 1.49 & \textbf{74.8} & 155  & \textbf{44.2} & 1173 \\
TF($k_{\max}{=}6$)          & 1.74 & 74.0 & 153  & 44.1 & 1192 \\
\midrule
TF(\textit{LowDiv})         & 1.13 & 72.3 & 291  & 40.7 & 1603 \\
\rowcolor{gray!8}
\textbf{TF(\textit{HighDiv})}
                            & 1.13 & \textbf{74.0} & 120  & \textbf{43.2} & 1056 \\
\bottomrule
\end{tabular}
\vspace{2pt}
\end{table}

\subsection{Effect of Error Supervision}

We study how the quantity and diversity of fused incorrect trajectories affect model
performance. Table~\ref{tab:combined-fusion-diversity} reports accuracy, the average
number of fused incorrect paths per training example (\textbf{\#Err}), and average
generated output length (\textbf{Tokens}) on GSM8K and MATH using LLaMA3-8B.

\paragraph{Effect of error quantity.}
Varying $k_{\max}$ controls the maximum number of incorrect trajectories that may be
included in a fused sequence. As shown in the top block of
Table~\ref{tab:combined-fusion-diversity}, increasing $k_{\max}$ from 1 to 4 raises both
\#Err and performance on both benchmarks. For instance, MATH accuracy improves from
39.56\% to 44.18\%, and GSM8K from 71.95\% to 74.83\%. These results suggest that a
moderate amount of error supervision provides additional informative signals beyond the
correct trajectory alone. However, increasing $k_{\max}$ further to 6 yields no
additional benefit despite producing more fused errors, indicating diminishing returns
when too many incorrect paths are included.

\paragraph{Effect of error diversity.}
We next control for the number of fused errors (\#Err=${}$1.138) and vary only the
selection strategy. \textbf{LowDiv} samples errors from the most frequent incorrect
answer bucket, whereas \textbf{HighDiv} prefers selecting trajectories from different
incorrect outcomes. As shown in the bottom block of
Table~\ref{tab:combined-fusion-diversity}, HighDiv consistently outperforms LowDiv
(+1.67\% on GSM8K and +2.50\% on MATH). HighDiv also produces substantially shorter
outputs. This indicates that redundant error patterns are less beneficial, while
structurally distinct incorrect trajectories provide more useful supervision and lead to
more concise reasoning behavior at inference time.

\subsection{Analysis}

\begin{figure}[t]
\centering
\includegraphics[width=\linewidth]{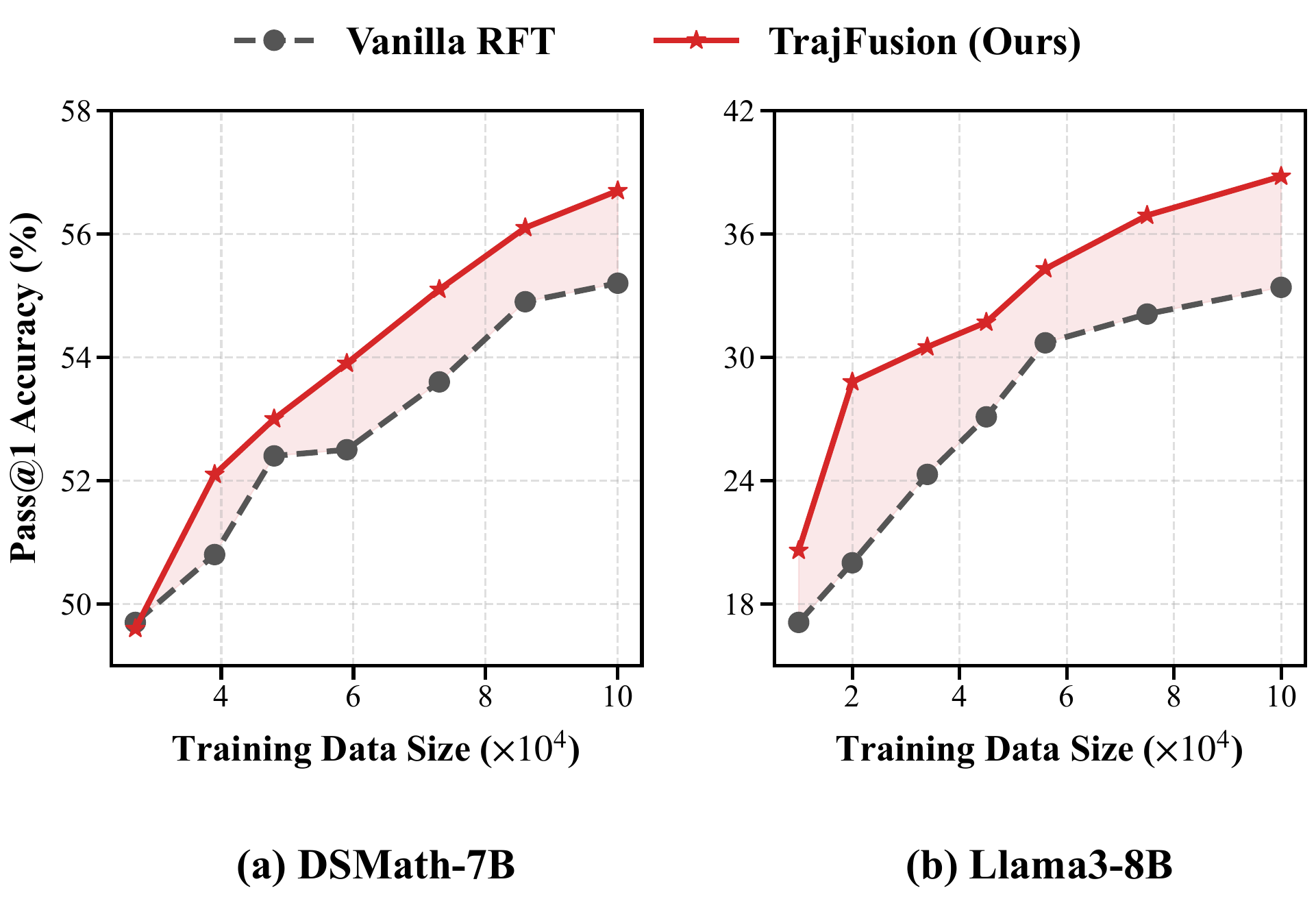}
\caption{
Pass@1 accuracy on MATH during the first training epoch, evaluated at fixed steps. 
TrajFusion (red) consistently outperforms Vanilla RFT (gray) across both models. 
The shaded areas highlight the performance margin, with a significant gap established early in training.
}
\label{fig:datasize-analysis}
\end{figure}

\paragraph{Data Efficiency.}
Figure~\ref{fig:datasize-analysis} shows the performance dynamics throughout the first epoch, evaluated at fixed intervals to track how accuracy changes with increasing data exposure.
In both DeepSeek-Math-7B and LLaMA3-8B model, TrajFusion consistently surpasses the Vanilla RFT baseline across all data intervals.
For LLaMA3-8B specifically, a distinct performance gap emerges as early as 20k samples and persists throughout the training process.
The shaded regions in the plots illustrate this robust margin, indicating that TrajFusion achieves higher accuracy than the baseline given the same number of training samples.
These observations suggest that the additional supervision from incorrect paths and reflections contributes to better sample efficiency.

\begin{figure}[t]
\centering
\includegraphics[width=\linewidth]{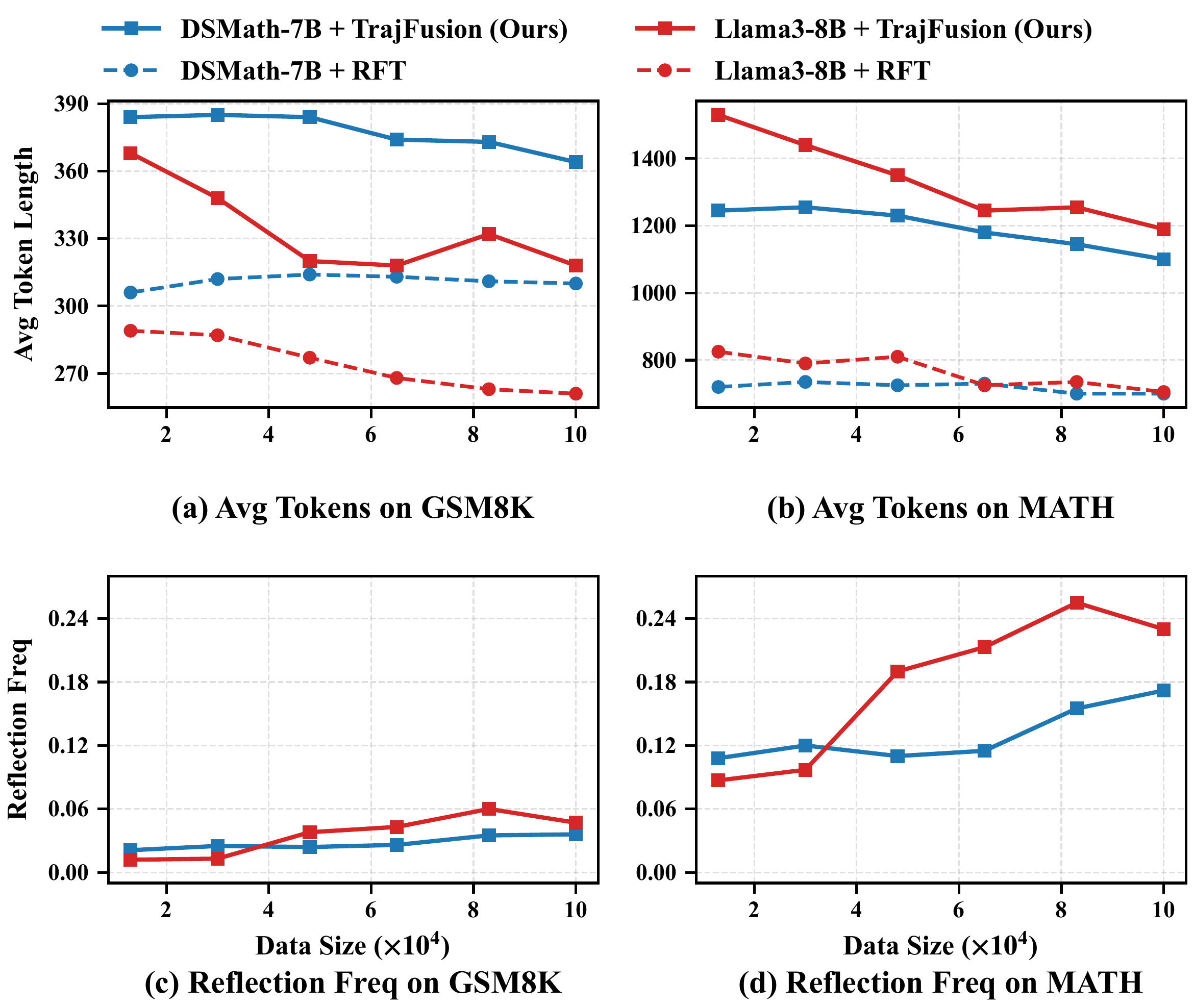}
\caption{
Average output length and reflection frequency on GSM8K and MATH.
\textbf{(a, b)} TrajFusion initially results in longer generations, but the average length gradually decreases as training progresses.
\textbf{(c, d)} The frequency of reflection tokens shows an upward trend, particularly on the more complex MATH dataset.
}
\label{fig:analysis}
\end{figure}

\paragraph{Token Efficiency and Reflection Behavior.}
We track the average output token length and reflection frequency to examine the generation behavior of TrajFusion (Figure~\ref{fig:analysis}).
Initially, TrajFusion models produce longer sequences than the baselines due to the inclusion of trial-and-error paths.
However, we observe a continuous decrease in average token count throughout the training process, suggesting that the generated trajectories become more concise.
Concurrently, the frequency of reflection tokens exhibits an upward trend.
This pattern is more distinct on the MATH dataset compared to GSM8K, which corresponds to the higher difficulty of the task.
The combination of decreasing total length and increasing reflection frequency implies that the model utilize the structural patterns more efficiently rather than merely producing longer sequences.

\subsection{Scaling to Long-form Reasoning Data}

\paragraph{Setup.}
Recent advances in mathematical reasoning have been facilitated by large reasoning models (LRMs)~\cite{deepseekai2025deepseekr1incentivizingreasoningcapability,moshkov2025aimo2, yang2025qwen3},
which train on long-form trajectories and operate with extended context windows. To examine
whether TrajFusion scales to this setting, we apply it to Qwen2.5-Math-7B-base, extending its context window from 4{,}096 tokens to 32{,}768 using
YaRN~\cite{peng2023yarn}.

We construct long-form training data from Deepseek-R1-Distill-Qwen-7B
\cite{deepseekai2025deepseekr1incentivizingreasoningcapability} and apply TrajFusion by
selecting diverse incorrect trajectories followed by a correct solution. Evaluation is
performed on three challenging benchmarks: AIME24, AIME25, and MATH-500. AIME contains competition-style problems with small validation splits; we report both
\textbf{Pass@1} and \textbf{Avg@16}. MATH-500~\cite{lightman2023let} is larger and evaluated using Pass@1 only.
To study the effect of context length, we also train 4K-context variants using data
sampled from Qwen2.5-Math-7B-Instruct and compare vanilla RFT with TrajFusion.

\begin{table}[tb]
\centering
\caption{
Performance of TrajFusion and vanilla RFT on AIME24, AIME25, and MATH-500 using
\textbf{Qwen2.5-Math-7B-base}. Results are reported for both 4K and 32K context.
}
\label{tab:fusion-longform}
\footnotesize
\setlength{\tabcolsep}{2pt}
\begin{tabular}{lccc}
\toprule
\textbf{Method} & \textbf{AIME24} & \textbf{AIME25} & \textbf{MATH-500} \\
\textit{} & \textit{P@1/Avg@16} &
             \textit{P@1/Avg@16} &
             \textit{P@1} \\
\midrule
RFT(4K)       & 16.7 / 15.6 & 10.0 / 12.1 & 79.4 \\
TrajFusion(4K) & \textbf{23.3} / 15.6 & \textbf{16.7} / \textbf{13.8} & \textbf{82.6} \\
\midrule
RFT(32K)      & 40.0 / 36.7 & 26.7 / 27.9 & 88.4 \\
TrajFusion(32K) & \textbf{43.3} / \textbf{40.8} &
                           \textbf{30.0} / \textbf{29.6} &
                           \textbf{90.4} \\
\bottomrule
\end{tabular}
\end{table}

\paragraph{Results.}
Table~\ref{tab:fusion-longform} shows that TrajFusion consistently outperforms vanilla RFT across all datasets and context lengths. 
In the 4K setting, TrajFusion improves Pass@1 from 16.7\% to 23.3\% on AIME24 and from 10.0\% to 16.7\% on AIME25. 
While the gains in Avg@16 are more moderate in this configuration, the substantial increase in Pass@1 suggests that TrajFusion effectively enhances the quality of top-ranked reasoning paths.

The performance advantage remains robust in the 32K setting utilizing long-form data. 
On AIME24, TrajFusion increases Pass@1 from 40.0\% to 43.3\% and Avg@16 from 36.7\% to 40.8\%. 
Similar improvements are observed on AIME25, and MATH-500 accuracy rises from 88.4\% to 90.4\%. 
These results indicate that TrajFusion scales effectively to extended context windows and complex reasoning trajectories.

\section{Related Work}
\subsection{Data Construction for Math Reasoning}

Prior work improves mathematical reasoning primarily by constructing
high-quality solution data.
Most approaches focus on expanding the coverage and diversity of
\emph{correct} reasoning trajectories, for example by modifying problem
difficulty, reformulating questions, recombining solution components,
or sampling multiple valid solution paths~\cite{luo2023wizardmath, Yu2023MetaMathQA, li2023mugglemath, Tong2024DART, guan2025rstar}.
To ensure reliability, these pipelines typically apply strict correctness
checks or rejection sampling, retaining only fully correct generations
as supervision~\cite{zelikman2022star, yuan2023scalingrelationshiplearningmathematical, xu2024chatglm, guan2025rstar, xiong2025minimalist}.
As a result, incorrect or partially incorrect trajectories are systematically
excluded at the data level.

\subsection{Negative Signals for Math Reasoning}

Several approaches introduce negative signals to improve
mathematical reasoning. Preference-based and reinforcement learning methods, such as
DPO~\cite{rafailov2023direct}, GRPO~\cite{shao2024deepseekmath}, and their variants~\cite{meng2024simpo, lu2024step, lai2024step},
leverage suboptimal or lower-reward samples to guide models away from
undesirable reasoning paths. NSR~\cite{zhu2025surprising} shows that exclusively optimizing negative samples in reinforcement learning yields significant gains. NFT~\cite{chen2025bridging} incorporates negative signals by modifying the
supervised training loss. Self-correction approaches~\cite{An2023LeMa, Pan2025LEMMA} introduce errors
through multi-turn data synthesis, exposing models to flawed trajectories and
their revisions. TrajFusion differs in that it operates within the RFT pipeline and directly
utilizes discarded samples. It does not require preference modeling or loss modification, enabling
single-stage learning from negative signals under standard next-token
prediction.

\section{Conclusion}
In this work, we proposed TrajFusion, a lightweight fine-tuning method that leverages both correct and incorrect reasoning trajectories during rejection sampling fine-tuning. By structuring diverse failed completions and reflection prompts into adaptive fused trajectories, TrajFusion provides structured supervision by a compact trail-and-error reasoning path without requiring architectural changes or modifications to the training objective.
Experiments across diverse math benchmarks and models show consistent gains over vanilla rejection fine-tuning, especially on challenging and long-form regimes. TrajFusion offers a simple yet effective way to scale mathematical reasoning through supervision.

\section{Limitations}

TrajFusion leverages supervision signals from problems where model sampling yields both correct and incorrect trajectories, as such cases provide the most informative reasoning variations.
Consequently, the strength of the supervision signal depends on the presence
of sufficient outcome diversity during sampling, while questions that are consistently solved correctly or incorrectly offer limited contrastive information, in which case TrajFusion naturally reduces to vanilla rejection sampling fine-tuning behavior.

\bibliography{custom}
\clearpage

\appendix

\section{Experimental Details}

\subsection{Sampling Setup}

We adopt a diverse sampling strategy to expose the model to both successful and unsuccessful reasoning paths.

\paragraph{Sampling Strategy.}
For each input query, we generate 16 independent completions using nucleus sampling (top-$p = 0.95$) with a temperature of 0.7. This configuration balances diversity and coherence, encouraging exploration of multiple reasoning trajectories without producing excessively random outputs.

\paragraph{Prompt Format.}
All completions are generated using a fixed instruction-style prompt: 
\begin{tcolorbox}[
    mypromptstyle, 
    width=\linewidth, 
    boxsep=2pt,       
    left=4pt, right=4pt, top=4pt, bottom=4pt
]
\small 
\texttt{Please reason step by step and put your answer within \textbackslash boxed\{\}.}
\end{tcolorbox}

This encourages models to follow a chain-of-thought reasoning process and produce a clearly marked final answer suitable for automatic extraction.

\paragraph{Output Filtering.}
To ensure semantically valid supervision signals, we filter generated responses to remove malformed outputs.
Specifically, we discard completions that:
\begin{itemize}
    \item exhibit excessive repetition;
    \item switch languages mid-response (e.g., mixing English and Chinese);
    \item lack meaningful reasoning steps or a final boxed answer.
\end{itemize}
This step ensures that the training data remains pedagogically useful.

\paragraph{Answer Extraction and Verification.}
We extract the final boxed or inline answer from each completion using simple pattern matching heuristics. To assess correctness, we apply the automatic math verifier from opencompass~\cite{2023opencompass}, which executes symbolic evaluation of the extracted answer against the ground truth. Each completion is labeled as \emph{correct} or \emph{incorrect} based on exact match criteria.

\paragraph{Model-Specific Configurations.}
Sampling is conducted using two teacher models: \textbf{Qwen2.5-Math-7B-Instruct} (4K context) and \textbf{DeepSeek-R1-Distill-Qwen-7B} (16K context). Each model uses its default maximum context length and the same sampling settings above. Queries for which no correct completions are found are excluded from training.

\subsection{Trajectory Fusion}

\paragraph{Construction.}
For each query, we identify the set of incorrect responses generated during rejection sampling. These responses are grouped by their final extracted values, and we select the shortest representative from each group to prioritize concise reasoning paths with minimal redundancy.

\paragraph{Sequence Formatting.}
A complete assistant reply is constructed by interleaving the selected incorrect responses with fixed reflective prompts, simulating a trial-and-error reasoning process. The sequence concludes with a correct solution followed by a positive feedback phrase and the final answer. The trajectory is encapsulated within a thinking block to provide structured supervision:

\begin{tcolorbox}[mypromptstyle, left=2pt, right=2pt]
\small
\texttt{<think>} \\
\texttt{[Incorrect Response 1]} \\
\texttt{[Reflective Comment]} \\
\texttt{\dots} \\
\texttt{[Correct Response]} \\
\texttt{[Success Phrase]} \\
\texttt{</think>} \\
\texttt{The answer is \textbackslash boxed\{[Answer]\}.}
\end{tcolorbox}

If no eligible incorrect completions are found, the sequence defaults to a standard format containing only the correct response within the \texttt{<think>} block, omitting the error and reflection components.

\subsection{Training Setup}

\paragraph{Framework and Resources.}
We perform training using the \texttt{ms-swift} framework with DeepSpeed ZeRO-1 optimization. All models are trained on 8$\times$A100 80GB GPUs using \texttt{bf16} mixed-precision and input sequence packing to improve efficiency.

\paragraph{Hyperparameters.}
Table~\ref{tab:training-model-comparison} provides the detailed training configurations for each base model. We evaluate models at fixed data intervals within the first epoch to monitor performance dynamics (Figure~\ref{fig:datasize-analysis}).

\begin{table*}[t]
\centering
\caption{Training configurations for different base models and data scales.}
\label{tab:training-model-comparison}
\renewcommand{\arraystretch}{1.2}
\setlength{\tabcolsep}{8pt}
\small
\begin{tabular}{lccccc}
\toprule
\textbf{Base Model} & \textbf{Teacher Model} & \textbf{Learning Rate} & \textbf{Length} & \textbf{Context} & \textbf{\#Samples} \\
\midrule
LLaMA3-8B-base & Qwen2.5-Math-7B-instruct & \texttt{5e-5} & 4096 & Standard & 15K/100K \\
DeepSeekMath-7B-base & Qwen2.5-Math-7B-instruct & \texttt{5e-5} & 4096 & Standard & 15K/100K \\
Qwen2.5-Math-7B-base & Qwen2.5-Math-7B-instruct & \texttt{2e-5} & 4096 & Standard & 15K \\
Qwen2.5-Math-7B-base & DeepSeek-R1-Distill-Qwen-7B & \texttt{2e-5} & 32768 & Extended & 100K \\
\bottomrule
\end{tabular}
\end{table*}

\section{Prompts and Templates}
\label{app:prompts}

\subsection{Dialogue Templates}
\label{app:dialogue_templates}

To ensure the models correctly interpret the structured reasoning tasks, we employ native dialogue formats for each model family. The following templates illustrate the precise placement of system instructions, user queries, and assistant responses.

\begin{tcolorbox}[mypromptstyle, title=Qwen2.5-Math Template]
\small
\begin{lstlisting}[basicstyle=\ttfamily\small, breaklines=true, columns=flexible, keepspaces=true]
<|im_start|>system
Please reason step by step, and put your final answer within \boxed{}.
<|im_end|>
<|im_start|>user
{input}
<|im_end|>
<|im_start|>assistant
{output}
\end{lstlisting}
\end{tcolorbox}

\begin{tcolorbox}[mypromptstyle, title=LLaMA3 Template]
\small
\begin{lstlisting}[basicstyle=\ttfamily\small, breaklines=true, columns=flexible, keepspaces=true]
<|begin_of_text|>
<|start_header_id|>system<|end_header_id|>
Please reason step by step, and put your final answer within \boxed{}.
<|eot_id|>
<|start_header_id|>user<|end_header_id|>
{input}
<|eot_id|>
<|start_header_id|>assistant
<|end_header_id|>
{output}
<|eot_id|>
\end{lstlisting}
\end{tcolorbox}

\begin{tcolorbox}[mypromptstyle, title=DeepSeek-Math Template]
\small
\begin{lstlisting}[basicstyle=\ttfamily\small, breaklines=true, columns=flexible, keepspaces=true]
User: {input}
Please reason step by step, and put your final answer within \boxed{}.

Assistant: {output}
\end{lstlisting}
\end{tcolorbox}

\subsection{Reflective Prompt Repository}
\label{app:reflective_prompts}

To internalize the trial-and-error reasoning process, TrajFusion utilizes a curated set of transition phrases. These prompts are inserted between reasoning trajectories to signify error recognition and subsequent correction. Table~\ref{tab:reflective_prompts} lists the complete repository of phrases used in our experiments.

\subsection{A Example of TrajFusion}

Below we provide a simple TrajFusion example.

\begin{exampleblock}{\textbf{Example of TrajFusion}}
\small

\textit{\textbf{Problem:} Determine $(r + s)(s + t)(t + r)$, given that $r,\,s,\,t$ are the real roots of $x^{3}+9x^{2}-9x-8$.}

\medskip
\textbf{\textit{TrajFusion Solution:}}

\textit{Given the polynomial $P(x) = x^3 + 9x^2 - 9x - 8$ with roots $r, s, t$, we need to determine the value of $(r+s)(s+t)(t+r)$. By Vieta's formulas, we know: $r + s + t = -9$, $rs + rt + st = -9$, $rst = 8$. We first try to expand the expression as $(r+s)(s+t)(t+r) = rs + rt + st + 2rst$, and substituting the values we get $-9 + 2 \cdot 8 = 7$.}

\textit{\textbf{Wait\dots{} that doesn't seem right. Let's try again.}}

\textit{Given the polynomial $P(x) = x^3 + 9x^2 - 9x - 8$ with roots $r, s, t$, we need to find the value of $(r+s)(s+t)(t+r)$. By Vieta's formulas, we know: $r + s + t = -9$, $rs + rt + st = -9$, $rst = 8$. We start by expressing $(r+s)(s+t)(t+r)$ in a different form: $(r+s)(s+t)(t+r) = (r+s+t)(rs+rt+st) - rst$. Substituting the values from Vieta's formulas, we get: $(-9)(-9) - 8 = 81 - 8 = 73$. Thus, the value of $(r+s)(s+t)(t+r)$ is $\boxed{73}$. \textbf{Ah, that makes sense now.}}

\normalsize
\end{exampleblock}

\begin{table*}[ht]
\centering
\caption{Complete repository of reflection and feedback phrases used in TrajFusion.}
\label{tab:reflective_prompts}
\begin{tcolorbox}[mypromptstyle, width=\textwidth, left=5pt, right=5pt]
\renewcommand{\arraystretch}{1.5}
\small
\begin{tabularx}{\textwidth}{l p{0.75\textwidth}}
\toprule
\textbf{Category} & \textbf{Candidate Phrases (Randomly Sampled)} \\
\midrule
\textbf{Error Recognition} & 
\texttt{"Wait... that doesn't seem right. Let's try again."} \newline 
\texttt{"Hmm, maybe we're missing something. Think carefully."} \newline 
\texttt{"That doesn't add up. Let's give it another go."} \newline 
\texttt{"Hold on. That might not be correct. Try again."} \\
\midrule
\textbf{Success Confirmation} & 
\texttt{"That makes sense now."} \newline \texttt{"Yes, this is correct."} \newline \texttt{"Ah, that makes sense now."} \\
\midrule
\textbf{Certainty (Optional)} & 
\texttt{"I'm absolutely certain this is correct."} \newline 
\texttt{"No doubt about it, this answer is right."} \newline 
\texttt{"I fully trust this result."} \\
\bottomrule
\end{tabularx}
\end{tcolorbox}
\end{table*}

\subsection{Evaluation Protocol}

\paragraph{Setup.}
All evaluations were conducted using the official Qwen2.5-Math toolkit, with inference executed through the same \texttt{vLLM} engine and prompt templates as used in training. Decoding was configured with temperature 0.6 and top-$p$ 0.95, encouraging diverse reasoning while maintaining output coherence.

\paragraph{Metrics.}
For competition-style datasets such as AIME24 and AIME25, we report both \textbf{Pass@1} and \textbf{Avg@16} using 16 sampled completions per query. For all other datasets, we evaluate using a single generation and report Pass@1. Table~\ref{tab:evaluation-metrics} summarizes these settings.

\begin{table}[ht]
\centering
\small
\caption{Evaluation Settings by Dataset}
\label{tab:evaluation-metrics}
\renewcommand{\arraystretch}{1.2}
\setlength{\tabcolsep}{4pt}
\begin{tabular}{lcc}
\toprule
\textbf{Dataset} & \textbf{Samples per Query} & \textbf{Reported Metrics} \\
\midrule
AIME24      & 16 & Pass@1, Avg@16 \\
AIME25      & 16 & Pass@1, Avg@16 \\
All others          & 1  & Pass@1 \\
\bottomrule
\end{tabular}
\end{table}

\paragraph{Benchmarks.}
We evaluate model performance across these mathematical reasoning benchmarks that vary in domain, difficulty, and format:

\begin{itemize}[leftmargin=1.5em]
    \item \textbf{GSM8K} contains 8,792 grade-school math word problems focused on multi-step arithmetic reasoning. Each question typically requires 2–8 steps to solve. We use the official test split of 1,319 examples.

    \item \textbf{MATH} is a competition-style dataset of 12,500 problems drawn from high school mathematics contests. It spans seven categories—ranging from algebra to geometry—and includes problems at five difficulty levels. We evaluate on the 5,000-problem test set.

    \item \textbf{CollegeMath} includes 2,818 curated problems sourced from undergraduate math textbooks. It covers topics such as algebra, calculus, probability, linear algebra, and differential equations, targeting college-level mathematical proficiency.

    \item \textbf{DeepMind Mathematics} consists of 1,000 short-form math problems designed to evaluate symbolic and numerical reasoning. The dataset covers core areas like arithmetic, algebra, calculus, and basic probability.

    \item \textbf{OlympiadBench-Math} is a benchmark of 675 Olympiad-style problems aimed at evaluating advanced mathematical reasoning. We use only the English text-only subset in our experiments.

    \item \textbf{TheoremQA} is a theorem-centric question-answering dataset built around 350 mathematical and scientific theorems. It includes 800 problems spanning domains such as mathematics, physics, computer science, and engineering, designed to test a model’s ability to apply domain-specific knowledge.

    \item \textbf{MATH‑500} is a carefully curated subset of 500 math problems drawn from diverse domains such as algebra, probability, geometry, and trigonometry. These problems are designed to assess a model’s ability to engage with structured, multi-step reasoning under compact evaluation settings.

    \item \textbf{AIME24} consists of 30 problems from the 2024 American Invitational Mathematics Examination, an upper-tier contest for top AMC performers. Each question demands intensive algebraic, geometric, or combinatorial reasoning, and requires an integer answer between 0 and 999.

    \item \textbf{AIME25} is the corresponding AIME-style test from 2025, offering a separate but structurally similar set of 30 high-difficulty problems to reduce potential overlap and memorization.
\end{itemize}

\section{Case Study}
We present some representative cases from the test set to qualitatively analyze the behavior of DeepSeek-Math-7B-TrajFusion.

\tcbset{
  promptbox/.style={
    colback=blue!2!white,
    colframe=blue!40!white,
    fonttitle=\bfseries,
    coltitle=white,
    sharp corners,
    boxrule=0.8pt,
    breakable,       
    width=\textwidth 
  }
}

\onecolumn

\begin{tcolorbox}[promptbox, title=A Case from TrajFusion-DSMath]
\small
\textbf{\textit{Problem:}} \textit{Simplify
\[
\frac{\sec x}{\sin x} - \frac{\sin x}{\cos x}.
\]}

\textbf{\textit{TrajFusion-DSMath Solution:}}

\texttt{<think>}
To simplify the expression \(\frac{\sec x}{\sin x} - \frac{\sin x}{\cos x}\), we start by expressing \(\sec x\) and \(\sin x\) in terms of \(\cos x\) and \(\sin x\):

\[
\sec x = \frac{1}{\cos x} \quad \text{and} \quad \sin x = \sin x.
\]

Substituting these into the expression, we get:

\[
\frac{\sec x}{\sin x} = \frac{\frac{1}{\cos x}}{\sin x} = \frac{1}{\cos x \sin x} \quad \text{and} \quad \frac{\sin x}{\cos x} = \frac{\sin x}{\cos x}.
\]

So the expression becomes:

\[
\frac{1}{\cos x \sin x} - \frac{\sin x}{\cos x}.
\]

To combine these fractions, we need a common denominator. The common denominator for \(\cos x \sin x\) and \(\cos x\) is \(\cos x \sin x\). Rewriting the second fraction with this common denominator, we get:

\[
\frac{\sin x}{\cos x} = \frac{\sin x \sin x}{\cos x \sin x} = \frac{\sin^2 x}{\cos x \sin x}.
\]

Now we can combine the fractions:

\[
\frac{1}{\cos x \sin x} - \frac{\sin^2 x}{\cos x \sin x} = \frac{1 - \sin^2 x}{\cos x \sin x}.
\]

We recognize that \(1 - \sin^2 x\) is a difference of squares, which can be factored as:

\[
1 - \sin^2 x = (\cos x + \sin x)(\cos x - \sin x).
\]

So the expression becomes:

\[
\frac{(\cos x + \sin x)(\cos x - \sin x)}{\cos x \sin x}.
\]

We can split this into two fractions:

\[
\frac{\cos x + \sin x}{\cos x} \cdot \frac{\cos x - \sin x}{\sin x}.
\]

This simplifies to:

\[
\left(1 + \frac{\sin x}{\cos x}\right) \left(\frac{\cos x}{\sin x} - 1\right).
\]

We know that \(\frac{\sin x}{\cos x} = \tan x\) and \(\frac{\cos x}{\sin x} = \cot x\), so the expression becomes:

\[
(1 + \tan x)(\cot x - 1).
\]

Using the identity \(\cot x = \frac{1}{\tan x}\), we can rewrite the expression as:

\[
(1 + \tan x)\left(\frac{1}{\tan x} - 1\right) = (1 + \tan x) \cdot \frac{1 - \tan x}{\tan x} = \frac{(1 + \tan x)(1 - \tan x)}{\tan x} = \frac{1 - \tan^2 x}{\tan x}.
\]

We know that \(1 - \tan^2 x = \sec^2 x - \tan^2 x = \sec^2 x - \frac{\sin^2 x}{\cos^2 x} = \frac{\cos^2 x}{\cos^2 x} - \frac{\sin^2 x}{\cos^2 x} = \frac{\cos^2 x - \sin^2 x}{\cos^2 x} = \frac{\cos 2x}{\cos^2 x}\), so the expression becomes:

\[
\frac{\frac{\cos 2x}{\cos^2 x}}{\tan x} = \frac{\cos 2x}{\cos^2 x} \cdot \frac{\cos x}{\sin x} = \frac{\cos 2x \cos x}{\cos^2 x \sin x} = \frac{\cos 2x}{\cos x \sin x} = \frac{2 \cos^2 x - 1}{\cos x \sin x} = \frac{2 \cos x}{\sin x} - \frac{1}{\cos x \sin x} = 2 \cot x - \csc x.
\]

Therefore, the simplified form of the expression is:

\[
\boxed{2 \cot x - \csc x}.
\]

\textbf{Wait... that doesn't seem right. Let's try again.}

To simplify the expression \(\frac{\sec x}{\sin x} - \frac{\sin x}{\cos x}\), we start by expressing \(\sec x\) and \(\sin x\) in terms of \(\cos x\) and \(\sin x\):

\[
\sec x = \frac{1}{\cos x} \quad \text{and} \quad \sin x = \sin x.
\]

Substituting these into the expression, we get:

\[
\frac{\sec x}{\sin x} = \frac{\frac{1}{\cos x}}{\sin x} = \frac{1}{\cos x \sin x} \quad \text{and} \quad \frac{\sin x}{\cos x} = \frac{\sin x}{\cos x}.
\]

So the expression becomes:

\[
\frac{1}{\cos x \sin x} - \frac{\sin x}{\cos x}.
\]

To combine these fractions, we need a common denominator. The common denominator for \(\cos x \sin x\) and \(\cos x\) is \(\cos x \sin x\). Rewriting the second fraction with this common denominator, we get:

\[
\frac{\sin x}{\cos x} = \frac{\sin x \sin x}{\cos x \sin x} = \frac{\sin^2 x}{\cos x \sin x}.
\]

Now we can combine the fractions:

\[
\frac{1}{\cos x \sin x} - \frac{\sin^2 x}{\cos x \sin x} = \frac{1 - \sin^2 x}{\cos x \sin x}.
\]

We know from the Pythagorean identity that \(1 - \sin^2 x = \cos^2 x\). Substituting this into the expression, we get:

\[
\frac{\cos^2 x}{\cos x \sin x} = \frac{\cos x}{\sin x}.
\]

The expression \(\frac{\cos x}{\sin x}\) is the definition of \(\cot x\). Therefore, the simplified form of the original expression is:

\[
\boxed{\cot x}.
\]

That makes sense now.

\texttt{</think>}The answer is \[\boxed{\cot x}.\]
\end{tcolorbox}

\begin{tcolorbox}[promptbox, title=A Case from TrajFusion-DSMath-7B]
\small
\textbf{\textit{Problem:}} \textit{If \(\frac{\sqrt{2x}}{\sqrt{3x - 1}} = \frac{3}{2}\), solve for \(x\). Express your answer in simplest fractional form.
}

\textbf{\textit{TrajFusion-DSMath Solution:}}

\texttt{<think>}
To solve the equation \(\frac{\sqrt{2x}}{\sqrt{3x-1}} = \frac{3}{2}\), we start by eliminating the square roots by squaring both sides of the equation:

\[
\left(\frac{\sqrt{2x}}{\sqrt{3x - 1}}\right)^2 = \left(\frac{3}{2}\right)^2
\]

This simplifies to:

\[
\frac{2x}{3x - 1} = \frac{9}{4}
\]

Next, we eliminate the fractions by cross-multiplying:

\[
4 \cdot 2x = 9 \cdot (3x - 1)
\]

This simplifies to:

\[
8x = 27x - 9
\]

To isolate \(x\), we subtract \(8x\) from both sides:

\[
0 = 19x - 9
\]

Then, we add 9 to both sides:

\[
9 = 19x
\]

Finally, we divide both sides by 19:

\[
x = \frac{9}{19}
\]

\textbf{Thus, the solution is:}
\[
\boxed{\frac{9}{19}}
\]

\texttt{</think>}

The answer is \[\boxed{\frac{9}{19}}\]

\end{tcolorbox}

\end{document}